%% file: MainPaper.tex
\documentclass{bmvc2k}
\usepackage{float}
\usepackage{color}
\usepackage{url}
\usepackage{hyperref}
\usepackage{geometry}
\usepackage{booktabs}
\usepackage{longtable}
\usepackage{multirow} 
\usepackage{colortbl}  
\usepackage{multicol}

\usepackage{wrapfig}
\usepackage{adjustbox}
\usepackage{makecell}

\usepackage{fancyhdr}
\fancypagestyle{bmvcsup}{%
  \fancyhf{}%
  \fancyhead[RO,LE]{\small BABAKHANI ET AL.: SUPPLEMENTARY MATERIAL}
  \fancyhead[LO,RE]{\small\bfseries \thepage}
  
}

\makeatletter
\let\ps@plain\ps@bmvcsup
\makeatother

\title{Deep Learning for Metabolic Rate Estimation from Biosignals: A Comparative Study of Architectures and Signal Selection}

\addauthor{Sarvenaz Babakhani}{sarvenaz.babakhani@ki.uni-stuttgart.de}{1}

\addauthor{David Remy}{david.remy@iams.uni-stuttgart.de}{2}
\addauthor{Alina Roitberg}{roitberg@uni-hildesheim.de}{3}

\addinstitution{
 Institute for Artificial Intelligence\\
 University of Stuttgart\\
 Stuttgart, Germany
}
\addinstitution{
 Institute for Adaptive Mechanical Systems\\
 University of Stuttgart\\
 Stuttgart, Germany
}
\addinstitution{
 Intelligent Assistive Systems Lab\\
 University of Hildesheim\\
 Hildesheim, Germany
}
\runninghead{Babakhani et al.}{Deep Learning for Metabolic Rate Estimation}

\def\etal{\emph{et al}\bmvaOneDot}

\begin{document}

\maketitle

\begin{abstract}

Energy expenditure estimation aims to infer human metabolic rate from physiological signals such as heart rate, respiration, or accelerometer, and has been studied primarily with classical regression methods. The few existing deep learning approaches rarely disentangle the role of neural architecture from that of signal choice. In this work, we systematically evaluate both aspects. We compare classical baselines with newer neural architectures across single signals, signal pairs, and grouped sensor inputs for diverse physical activities. Our results show that minute ventilation is the most predictive individual signal, with a transformer model achieving the lowest root mean square error (RMSE) of $0.87 \,\mathrm{W/kg}$ across all activities. Paired and grouped signals, such as those from the Hexoskin smart shirt (5 signals), offer good alternatives for faster models like CNN and ResNet with attention.
Per-activity evaluation revealed mixed outcomes: notably better outcomes in low-intensity activities (RMSE down to $0.29 \,\mathrm{W/kg}$; NRMSE = $0.04$), while higher-intensity tasks showed larger RMSE but more comparable normalized errors.
Finally, subject-level analysis highlights strong inter-individual variability, motivating the need for adaptive modeling strategies. Our code and models will be publicly available at \href{https://github.com/Sarvibabakhani/deeplearning-biosignals-ee}{this GitHub repository}.

\end{abstract}

\section{Introduction and Related Work}
\label{sec:intro}
Wearable assistive devices are promising for improving mobility, optimizing body energy expenditure, and enhancing the quality of life for older adults and individuals with mobility impairments~\cite{K.A.Ingraham}. Designing such systems is challenging due to the complexity of the human neuromuscular system. To address this, human- and body-in-the-loop optimization methods adapt device parameters in real time based on user feedback, thereby reducing reliance on complete biomechanical models and avoiding manual tuning in clinical settings \cite{P.Slade, koller2016body, felt2015body}.

A critical component of such body-in-the-loop optimization systems is the accurate estimation of energy expenditure (EE). One precise but intrusive way for measuring EE is indirect calorimetry, which requires the measurement of oxygen consumption (\(\dot{V}_{O_{2}}\)) and carbon dioxide production (\(\dot{V}_{CO_{2}}\)) via a metabolic mask; thus limiting long-term use.  As an alternative, portable wearable sensors can provide physiological signals, such as heart rate, respiration, or accelerometry, which can be combined to estimate energy expenditure. To achieve this estimation, classical ML methods were used largely in earlier studies, often with limited sensor modalities and task types.
\begin{figure}
\centering
\includegraphics[width=\textwidth]{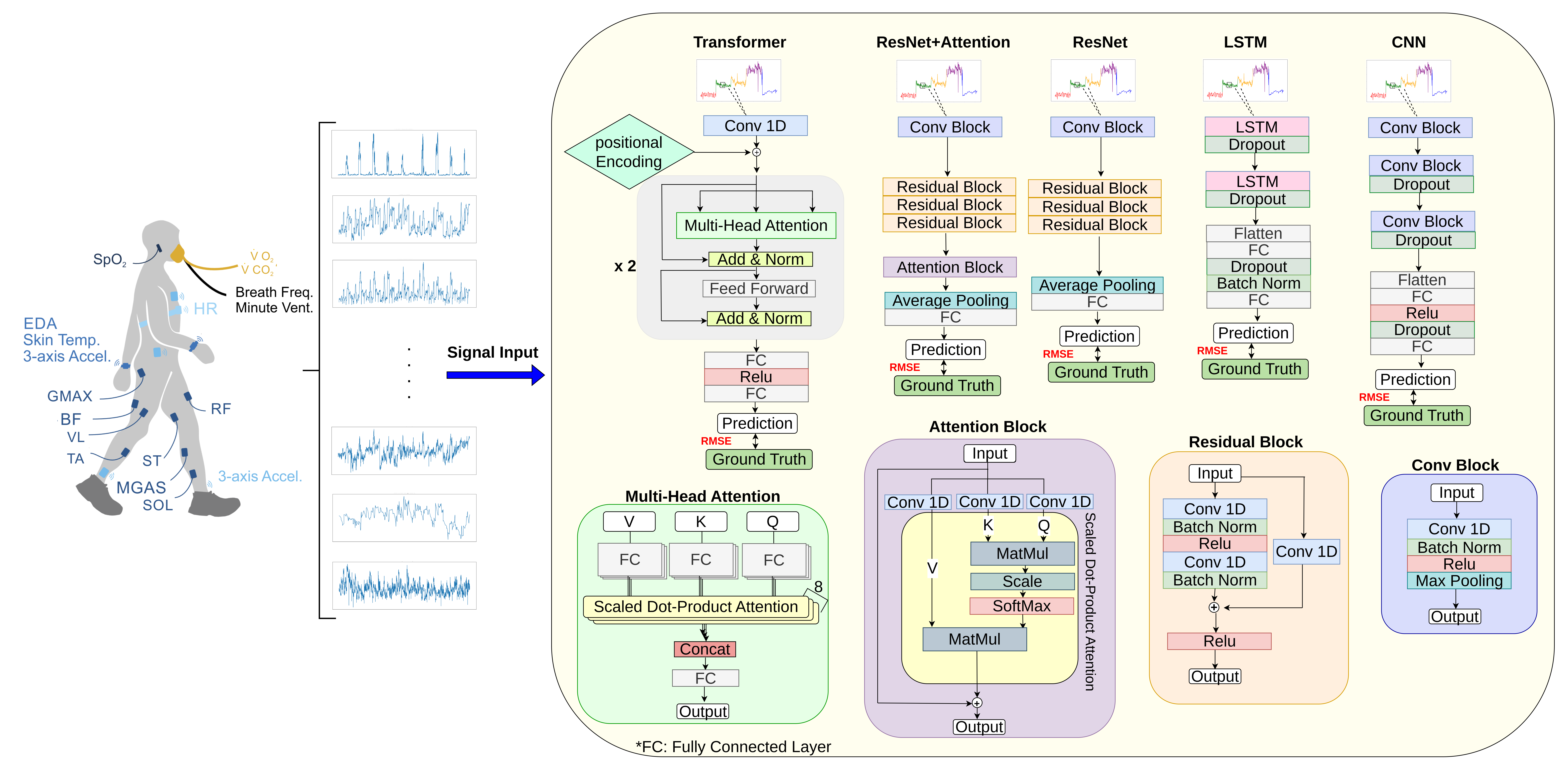}
\caption{Multimodal physiological signal processing pipeline for  EE. Wearable sensors placed across the body collect multimodal signals. These signals are processed and fed as input into multiple neural network architectures. (Image of sensor placement on the body is adapted from Ingraham \etal \cite{K.A.Ingraham}).}
\label{whole_pic}
\end{figure}
For instance, \cite{Falcone, Marena, Monteiro2,monteiro2024hitl,Slade2} applied linear regression, SVMs, or Gaussian Process Regression with a limited number of signals like accelerometry and heart rate. Some of them reported errors as low as $0.31\text{--}0.66 \,\mathrm{W/kg}$, but the number of activities was restricted.
Other studies, such as Cvetković \etal \cite{Cvetković} attempted richer sensor sets, but still optimized only for narrow activity ranges.
Ingraham \etal \cite{K.A.Ingraham},  evaluated the importance of physiological signals for EE estimation and released an accompanying public dataset. They utilized linear regression models and indicated that using minute ventilation alone achieved an RMSE of $1.24 \,\mathrm{W/kg}$. While their dataset has since enabled broader evaluation, their analysis was limited to ML approaches. In contrast, our work leverages the same dataset to benchmark both ML and Deep Learning methods across a wide range of signals and activities.

 Deep learning studies have achieved stronger performance but under narrower conditions. 
In vision-based approaches for estimate metabolic, neural architectures (CNNs and Transformers)~\cite{masullo2018calorinet,peng2022should,kasturi2024estimating} are a common choice, but they provide only coarse estimates. In contrast, wearable-sensor research has seen less widespread use of deep learning, though also here several studies exist with promises of better estimation through such architectures.
 For example, a Deep Multi-Branch Two-Stage Regression Network (DMTRN)  was introduced by Ni \etal \cite{Ni} that utilized ECG and IMU data, and achieved an RMSE of $0.71 \,\mathrm{kcal/min}$. Other studies, such as those by Lopes \etal \cite{Lopes}, Lee and Lee \cite{lee}, and Yuan \etal \cite{Yuan}  employed CNNs, LSTMs, or hybrid CNN–LSTM models on signals such as IMU, EMG, and motion velocity, but activity diversity in their studies was focused on walking-based tasks, which reduces generalizability. In parallel, Kim and Seong \cite{kim} introduced a personalized EE estimation method that combines a modified MET formulation with a heart rate–driven Deep Q-Network, achieving improved per-subject accuracy but without demonstrating cross-subject generalization. Other research has explored specific scenarios, further limiting broader applicability. For instance, \cite{Slade3}  estimated EE during assisted and loaded walking, reaching RMSE values as low as $0.40 \,\mathrm{W/kg}$ across novel subjects and conditions, while \cite{houssein2023energy} evaluated model performance separately for each activity, without assessing generalization across all activities.
Additionally, \cite{xu2024spatial} introduced a spatial-temporal fusion network with hybrid attention mechanisms, using multi-sensor data (sEMG, IMU, and HR). Results indicated strong performance with an RMSE of $0.342 \,\mathrm{kcal/min}$ in individual scenarios and a cross-subject RMSE of $0.646 \,\mathrm{kcal/min}$. 
However, the study did not address the average performance across all subjects or scenarios, making it difficult to assess overall generalization. Our study extends this line of work by considering a broader sensor set, diverse activities, and a wider variety of architectures beyond a single model class.

In summary, most prior work on EE estimation has relied on classical machine learning techniques, while only a few recent studies have explored deep learning. However, these deep learning approaches rarely disentangle the role of neural architecture from that of signal selection, leaving open the question of architectural and signal choices. 
In this study, we expanded on recent research by comparing various models, including linear regression, CNN, ResNet, ResNet+Attention, LSTM, and Transformer models, across multiple input configurations (single, paired, and grouped signals) to check both overall activity and activity-specific performance. We studied the ability of the models to generalize and investigated the impact of transitions between activities. Additionally, we examined inter-individual variability in signal effectiveness and model performance by studying a per-subject evaluation.
By systematically comparing classical and deep learning methods on the Ingraham \etal \cite{K.A.Ingraham} dataset, we establish a new state of the art on this benchmark, achieving substantially lower error rates than prior work.

\section{Methods}
\subsection{Dataset}

In this study, we use the public dataset provided by Ingraham \etal 
Please refer to \cite{K.A.Ingraham} for more information on data collection and processing.
In total, sixteen signals were gathered with wearable sensors from 10 different subjects, performing six types of physical activities. 
The signals are provided in Table~\ref{tab:model-performance}, with additional details available in the supplementary material. The ground truth energy expenditure was computed using the Brockway equation~\cite{Brockway} and normalized based on the subject's body weight. 

\noindent\textbf{Grouping Signals:} In addition to evaluating models on 1) individual signals, we considered 2) all possible signal pairs as well as 3) physiologically motivated signal groups proposed by \cite{K.A.Ingraham}. Group memberships are listed in Table~\ref{tab:model-performance}, with each signal annotated by its group label in parentheses (e.g., G for Global signals).  \textit{Global signals}, such as minute ventilation and heart rate, reflect whole-body physiological state. 
\textit{Local signals}, such as ankle and wrist acceleration,  capture activity in specific body segments. 
The \textit {Local+Global} setting combines both groups, incorporating all 16 signals. \textit{Hexoskin signals} refer to those measured using the Hexoskin smart shirt \cite{hexo}, like breath frequency and chest acceleration.

\subsection{Neural Architectures}\label{parameter}

We design and analyze different neural network-based models for estimating human metabolic rate from wearable sensor signals, aiming to disentangle the impact of neural architecture from that of signal choice by comparing models with distinct inductive biases.
We consider six representative models: Linear Regression, CNN, LSTM, ResNet, ResNet+Attention, and Transformer.
All deep learning models in this study operate directly on temporal signal inputs.
An overview of the task and implemented approaches is given in Figure \ref{whole_pic}.\\
\textbf{Linear Regression:}
This simple and interpretable model serves as an important baseline, as it is widely used in prior work on EE estimation, and is the key approach used in the benchmark we build on~\cite{K.A.Ingraham}. We implemented both single and multiple linear regression variants. \\
\textbf{CNN:}
This model is designed to capture local temporal patterns in the input signals by applying one-dimensional convolutions through time.
The model consists of three 1D convolutional blocks followed by fully connected layers.
The convolutional output is flattened and passed through two fully connected layers.
Finally, the output layer has a linear activation function to match the prediction with the target dimensions.
(The training time is $550.98 \,\mathrm{s}$). \\
\textbf{LSTM:} We implemented a stacked LSTM-based regression network to leverage both short-term and long-term memory to monitor changes in the input signals over time. The model consists of two sequential Long Short-Term Memory (LSTM) layers. The final LSTM output is flattened and passed through a fully connected layer, followed by batch normalization and dropout. (The training time is $ 264.06 \,\mathrm{s}$).\\
\textbf{ResNet:}
We build on the popular  ResNet architecture \cite{resi} and adapt it for 1D time-series input. The main idea is to utilize residual (skip) connections to allow the network to pass in-

\begin{table}[ht]
\centering
\resizebox{\textwidth}{!}{%
\begin{scriptsize} 
\begin{tabular}{@{}lccccccc@{}}
\toprule
\textbf{Signal} & \textbf{\makecell{Lin-Reg \\\cite{K.A.Ingraham}}}&\textbf{\makecell{Lin-Reg \\ (ours)}}  & \textbf{CNN} & \textbf{LSTM} & \textbf{ResNet} & \textbf{ResNet+Att} & \textbf{Transformer} \\
\midrule
Waist Acceleration (L,H)         & - & 2.41 & 2.22 & 2.04 & 2.30 & 3.04 & \textbf{1.89} \\
Chest Acceleration (L,H)         & -& 2.35 & 2.01 & 2.02 & 2.09 & 2.01 & \textbf{1.92} \\
Left Ankle Acceleration (L)      & -& 2.33 & \textbf{1.84} & 2.02 & 2.01 & 1.89 & 1.85 \\
Right Ankle Acceleration (L)    & - & 2.33 & \textbf{1.83} & 1.96 & 1.92 & 1.87 & 1.83 \\
Left Wrist Acceleration (L)      & -& 2.70 & 2.17 & 2.16 & 2.19 & \textbf{2.08} & 2.09 \\
Left Wrist Electrodermal (G)     &2.93& 3.19 & 2.60 & 2.36 & 2.53 & 2.83 & \textbf{2.11} \\
Left Wrist Temperature (G)       & -& 3.11 & 2.55 & 2.81 & 2.73 & 2.54 & \textbf{2.52} \\
Right Wrist Acceleration (L)     & -& 2.73 & 2.16 & 2.22 & 2.25 & 2.25 & \textbf{2.07} \\
Right Wrist Electrodermal (G)    & -& 3.01 & 2.37 & 2.46 & 2.59 & 2.88 & \textbf{2.24} \\
Right Wrist Temperature (G)     & - & 3.13 & \textbf{2.53} & 2.74 & 2.85 & 2.56 & 2.56 \\
EMG Magnitude Left (L)          & - & 2.86 & 2.40 & 2.48 & 2.58 & \textbf{2.37} & 2.40 \\
EMG Magnitude Right (L)         & - & 2.83 & \textbf{2.40} & 2.55 & 2.50 & 2.43 & 2.42 \\
Heart Rate (G,H)                & - & 2.29 & \textbf{1.81} & 2.08 & 1.97 & 1.84 & 1.95 \\
$SpO_{2}$ (G)             & - & 2.81 & 2.34 & 2.51 & 2.48 & \textbf{2.33} & 2.34 \\
Breath Frequency (G,H)         & -  & 2.89 & 2.46 & 2.67 & 2.57 & \textbf{2.35} & 2.43 \\
\rowcolor{gray!20}
Minute Ventilation (G,H)         &1.24& 1.30 & 1.00 & 1.03 & 1.03 & 0.97 & \textbf{0.87} \\
\midrule
Global Signals                  &1.25 & 1.34 & \textbf{0.97} & 1.08 & 1.16 & 1.17 & 1.18 \\
Global Signals W/O MinVent          & -& 2.35 & 1.82 & \textbf{1.77} & 1.96 & 1.81 & 2.17 \\
Local Signals                  & -  & 1.99 & 1.98 & 1.73 & 1.84 & 2.69 & \textbf{1.54} \\
Local+Global Signals             &1.28& 1.27 & \textbf{0.93} & 1.13 & 1.34 & 1.21 & 1.27 \\
Local+Global W/O MinVent         & -  & 1.88 & \textbf{1.60} & 1.79 & 1.95 & 1.88 & 1.58 \\
Hexoskin Signals                 &1.24& 1.28 & \textbf{0.92} & 0.98 & 1.12 & 1.10 & 1.07 \\
\midrule
$\dot{V}_{O_{2}}$ (part of ground truth)                &0.93& 0.91 & 0.62 & 0.56 & 0.58 & 0.74 & \textbf{0.40} \\
\bottomrule
\end{tabular}
\end{scriptsize}
}
\caption{RMSE (W/kg) of Models using physiological input signals (individual and grouped). Signal grouping's initials: (L) local, (G) Global, and (H) Hexoskin. The first column reports baseline (linear regression) results from \cite{K.A.Ingraham}(where available), while the remaining columns present our reproduced linear regression and deep learning models.}
\label{tab:model-performance}
\end{table}
\noindent formation from earlier layers directly to later layers.
The model architecture begins with a 1D convolutional block.
Next, there are three residual blocks with increasing output dimensions. Global average pooling is applied after the last residual block, followed by a linear layer mapping to the output size.(The training time is   $ 227.01 \,\mathrm{s}$).\\
\textbf{ResNet+Attention:}
We extend the ResNet architecture with a self-attention block after the residual layers to capture longer-range dependencies  (see Figure~\ref{whole_pic}).
This block uses a self-attention mechanism over the temporal dimension. In this block, there are three separate $1\times1$ convolutions to produce query, key, and value, which represent the input. The attention score is calculated with the related equation for $Attention (Q, K, V)$ in \cite{vaswani}. 
After re-weighting the values, a residual connection adds the attention output back to the input and preserves the original features. (The training time is $ 590.99 \,\mathrm{s}$).\\ 
\textbf{Transformer:}
To model complex temporal dependencies that extend beyond local patterns, we implemented a multi-head attention model as a Transformer-based architecture inspired by the original Transformer encoder framework \cite{vaswani}. This model combines 1D convolutional feature projection, positional encoding, and multi-head self-attention.
A stack of two Transformer encoder layers is applied, each consisting of multi-head self-attention (with eight heads), feedforward network, residual connections, and layer normalization. Finally, the output is passed through a small feedforward network to match the output dimension of the EE prediction.(The training time is  $4877.41 \,\mathrm{s}$).\\
Main experiments were conducted using PyTorch $2.5.1$ with CUDA $12.4$ on a single NVIDIA GeForce RTX $4090$ GPU with $24$ GB memory; additional hyperparameters and training configurations are provided in the supplementary material.
\section{Experiments and Result}

We follow the preprocessing and evaluation protocol of ~\cite{K.A.Ingraham} and evaluate the architectures using leave-one-subject-out cross-validation. In each fold, one subject is used for testing, while the remaining subjects are used for training, with $15\%$ of the training data held out for validation. This process is repeated for all 10 subjects. For each test subject, root mean square error (RMSE) is computed as the average error across all predicted time steps, and the final RMSE is obtained by averaging the results across all folds. When analyzing performance for each activity, RMSE is normalized by the average EE of that activity to obtain the normalized RMSE (NRMSE).\\
A manual grid search over hyperparameters is performed using minute ventilation as the input signal, and the selected values are then fixed for training on all other signals.
\subsection{Single and Grouped signals Comparison}

We began by examining model performance across single and grouped signals in Table \ref{tab:model-performance}.
Minute ventilation emerged as the most reliable predictor of EE, consistently outperforming other modalities.
Among models, the Transformer-based approach achieved the lowest overall error (RMSE of $0.87  \,\mathrm{W/kg}$ for minute ventilation), and was also best for several other signals, such as waist and chest acceleration. 
If we prefer an alternative for minute ventilation (challenging to measure, see Sec. \ref{Alter}), heart rate is a viable alternative due to its ease of measurement and the second lowest RMSE ($1.81  \,\mathrm{W/kg}$) among other signals.
Beyond the transformer, other architectures also showed strengths: the ResNet+Attention model outperformed on four signals, whereas the CNN attained the lowest error on five signals and was particularly effective on grouped inputs.
Recognition quality was high for Hexoskin signals (RMSE of $0.92  \,\mathrm{W/kg} $), similar to the Transformer's best result on minute ventilation.

\subsection{Pair Combination of Signals}
\begin{figure}[hbpt]
\centering

\begin{minipage}[b]{0.48\textwidth}
    \centering
    \includegraphics[width=\textwidth]{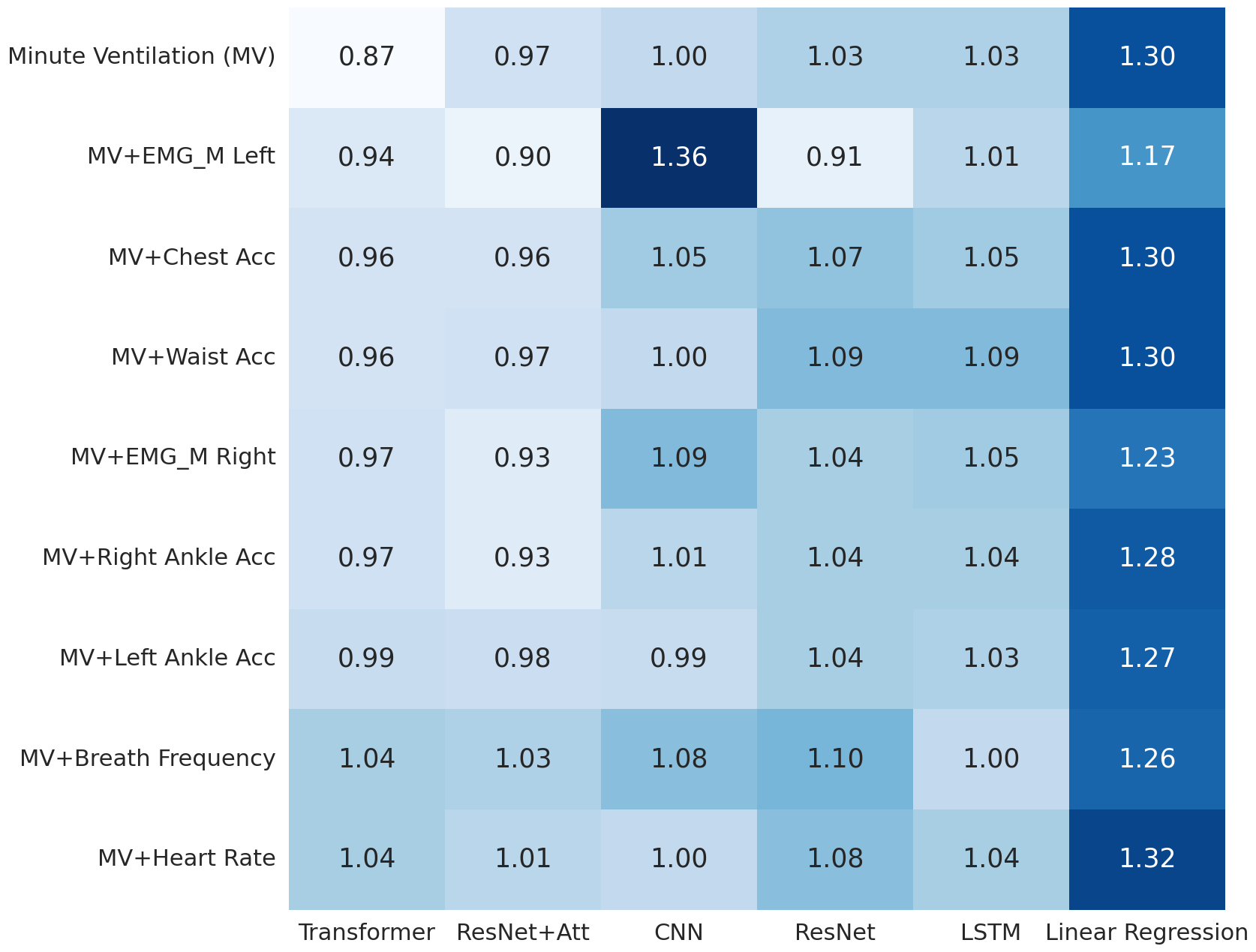}
    
    \caption{Heatmap of RMSE values using Minute Ventilation (MV) alone and in combination with secondary signals (rows). The columns correspond to different prediction models. Lower RMSE values (lighter colors) indicate better predictive performance.}
    \label{fig:heatmap}
\end{minipage}
\hfill
\begin{minipage}[b]{0.48\textwidth}
    \centering
    \includegraphics[width=\textwidth]{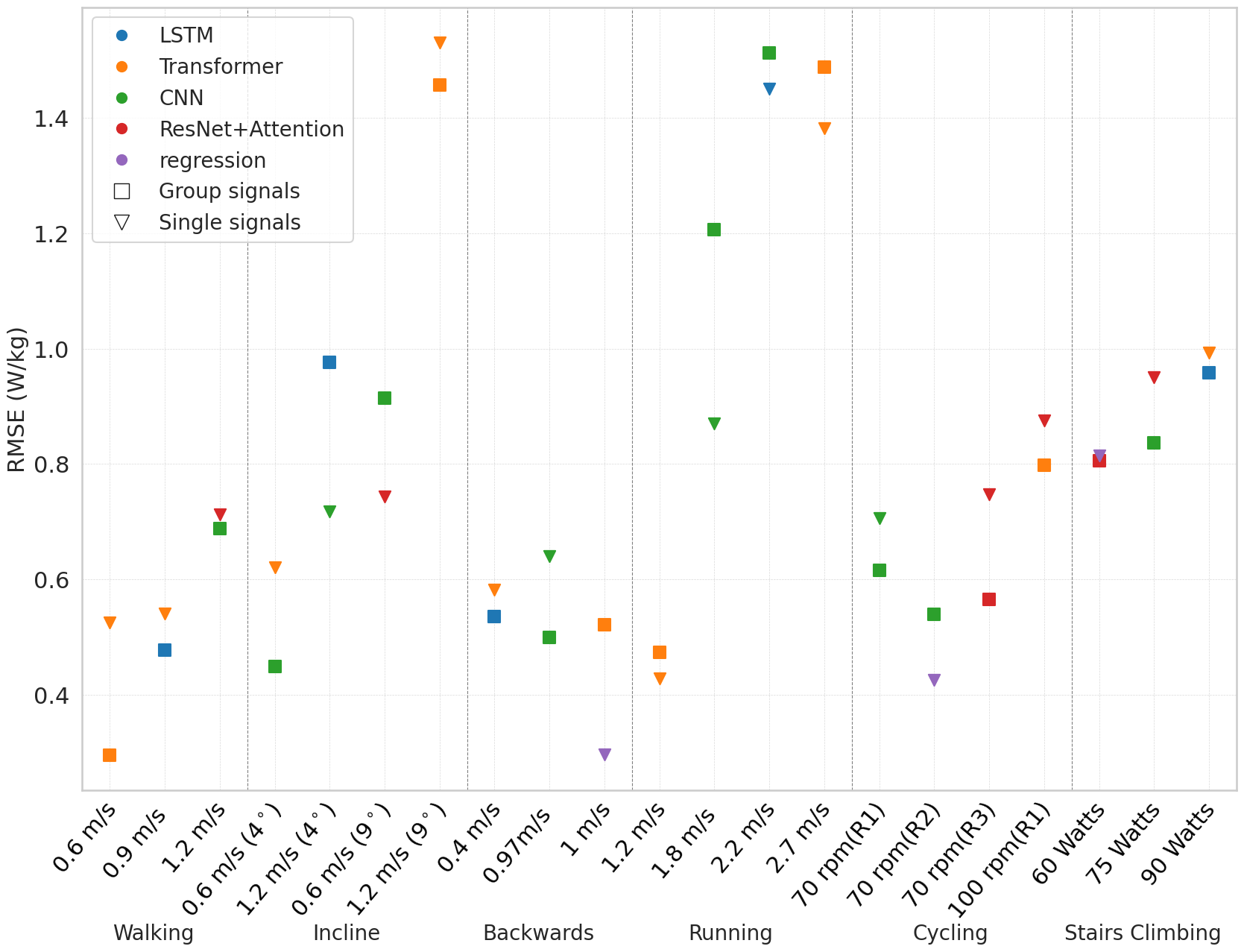}
    
    \caption{Model performance across different activities and conditions. The x-axis shows six activities with variations in speed or resistance.  Model types are distinguished by color, while input type (single and grouped) is indicated by marker shape.}
    \label{fig:peractiv}
\end{minipage}

\vspace{0.4em}

\label{fig:sidebyside}
\end{figure}

In the next step, we considered pairwise combinations of physiological signals.
While individual signals alone may carry strong predictive power, combining their sources can reveal useful synergistic effects.
As expected, the combination of minute ventilation and other signals consistently outperformed all other combinations. Among these combinations, we selected the best ones and visualized them in Figure \ref{fig:heatmap}. 
The first row showed the result when minute ventilation is the only input, and the other rows illustrated the results when additional signals were combined with minute ventilation. 
Both the Transformer and ResNet+Attention yielded the best overall results.
Notably, adding EMG magnitude (left) further boosted the performance of both ResNet+Attention and ResNet. 
Beyond accuracy, both networks benefited from faster training times compared to the Transformer (see Sec. \ref{parameter}). When compared to the results in Table \ref{tab:model-performance}, pairing signals enabled ResNet+Attention and ResNet to surpass the CNN with Hexoskin inputs (RMSE $0.92 \,\mathrm{W/kg}$), underscoring the added value of EMG signals in combination with minute ventilation.

\subsection{Alternatives to Minute Ventilation} \label{Alter}

While minute ventilation is the strongest predictor in our study, its measurement is technically demanding, costly, and often uncomfortable, as it typically requires the use of a mask. This motivated the search for practical alternatives.

\begin{figure}[hbpt]
\centering
\includegraphics[width=\textwidth]{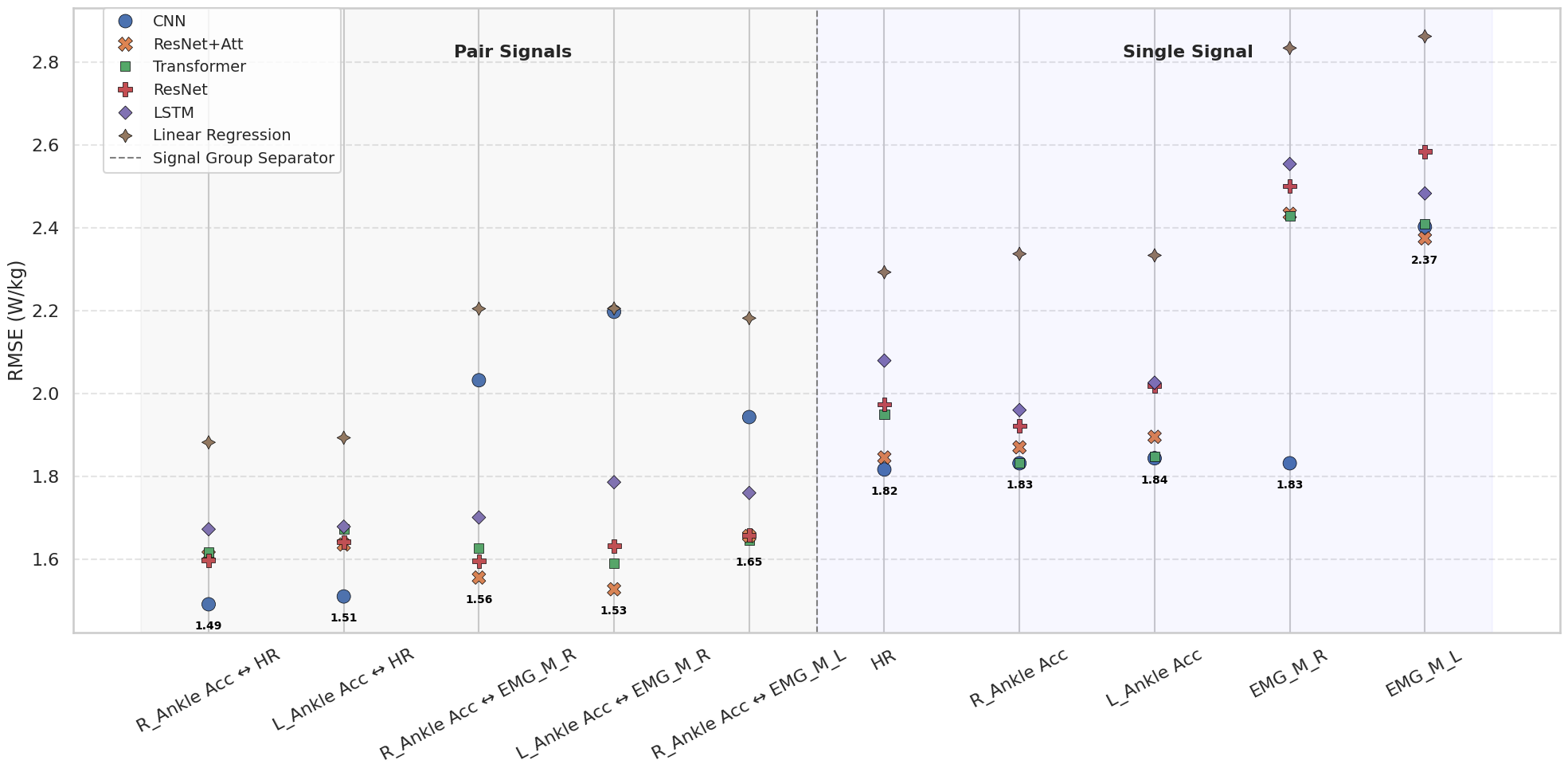}
\caption{Model performance using alternative input signals for Minute Ventilation. The left panel shows results from paired signal combinations, while the right panel shows single-signal inputs.  Different models are represented by distinct colors and markers. }
\label{fig_NotMin}
\end{figure}

In Figure \ref{fig_NotMin}, we compared five candidate physiological signals beyond minute ventilation. The right side of the figure shows their individual performance, while the left side highlights the best-performing pairwise combinations.
Pairing signals resulted in lower RMSE, indicating that when minute ventilation is removed, using other signals in pairs is more beneficial.

As we mentioned before, CNN with heart rate was the best single signal after minute ventilation, but using it with the (right and left) ankle acceleration (RMSE:  $1.49$ and $ 1.51 \,\mathrm{W/kg} $) improved the performance by almost $17\%$.
Another effective pair was left ankle acceleration and EMG magnitude (right) with the ResNet+Attention, yielding an RMSE of $1.53 \,\mathrm{W/kg} $.

When analyzing the most effective pair for each signal (in the absence of minute ventilation), heart rate and ankle acceleration (left or right) frequently emerged as the strongest partners. Across the majority of best pairs cases, CNN delivered the best predictive performance, achieving the lowest RMSE relative to other models.

While CNN was effective for certain signal combinations (grouped or pairs), poor signal selection led to significantly worse results. For example, pairing EMG with electrodermal activity produced the highest RMSE ($8.05 \,\mathrm{W/kg} $), underscoring the poor suitability of these signals for this task. Similarly, electrodermal and temperature signals, whether considered individually or in pairs, consistently yielded high errors (e.g., $3.15\text{--}3.23 \,\mathrm{W/kg} $), across diverse models, highlighting their limited predictive value.

The complete tables for the best partner(pair) of each signal and for the least effective pairs, along with the corresponding models, are provided in the supplementary materials.
\subsection{Per-Activity Evaluation}

Next, we evaluated model performance per activity, including different speeds and resistance conditions. The models were trained on all activity types, as in the previous experiments, but testing was carried out separately for each activity.

We observed two central findings. First, Figure \ref{fig:peractiv} showed that  Transformer and CNN performed best with single inputs. Linear regression with minute ventilation achieved the lowest single-signal RMSE of $0.29  \,\mathrm{W/kg} $ in the backward walking at $1 \,\mathrm{m/s} $. Grouped signals with CNN-, LSTM-, and Transformer-based methods consistently improved over single inputs. The best overall result, also $0.29  \,\mathrm{W/kg} $, was obtained by the Transformer with Local+Global data during walking at $0.6 \,\mathrm{m/s} $.
\begin{table}[hbpt]
\centering

\begin{minipage}{0.47\textwidth}
\centering

\scalebox{0.45}{ 
  \renewcommand{\arraystretch}{1.2}
\begin{tabular}{@{}lcccccc@{}}
\toprule
Activity & Condition & NRMSE\_single & Signal & NRMSE\_group & group \\
\midrule
\multirow{3}{*}{Walking} 
& 0.6 m/s  & 0.14 & Min\_Vent & 0.08 & Loc+Glob \\
& 0.9 m/s  & 0.13 & Min\_Vent & 0.11 & Loc+Glob \\
& 1.2 m/s  & 0.14 & Min\_Vent & 0.14 & Loc+Glob \\
\midrule
\multirow{4}{*}{Incline} 
& 0.6 m/s ($4^{\circ}$)  & 0.13 & Min\_Vent & 0.09 & Loc+Glob \\
& 1.2 m/s ($4^{\circ}$)  & 0.09 & L\_Wrist\_Elec    & 0.13 & Hexoskin \\
& 0.6 m/s ($9^{\circ}$)  & 0.10 & Min\_Vent & 0.13 & Hexoskin \\
& 1.2 m/s ($9^{\circ}$)  & 0.12 & Min\_Vent & 0.11 & Hexoskin \\
\midrule
\multirow{3}{*}{Backwards} 
& 0.4 m/s  & 0.15 & Min\_Vent & 0.14 & Loc+Glob \\
& 0.7 m/s  & 0.13 & Min\_Vent & 0.10 & Loc+Glob \\
& 1.0 m/s  & \textbf{0.04} & Min\_Vent & 0.08 & Hexoskin \\
\bottomrule
\end{tabular}%
}
\end{minipage}
\hfill
\begin{minipage}{0.5\textwidth}
\centering
\scalebox{0.43}{ 
  \renewcommand{\arraystretch}{1.18}
\begin{tabular}{@{}lcccccc@{}}
\toprule
Activity & Condition & NRMSE\_single & Signal & NRMSE\_group & group \\
\midrule
\multirow{4}{*}{Running} 
& 1.2 m/s  & 0.09 & Min\_Vent & 0.10 & Global \\
& 1.8 m/s  & 0.08 & Min\_Vent & 0.11 & Hexoskin \\
& 2.2 m/s  & 0.12 & R\_Ankle\_ACCL    & 0.12 & Hexoskin \\
& 2.7 m/s  & 0.09 & Min\_Vent & 0.10 & Local \\
\midrule
\multirow{4}{*}{Cycling} 
& 70 rpm (R1)  & 0.13 & Chest\_ACC  & 0.11 & Global \\
& 70 rpm (R3)  & 0.06 & $SpO_{2}$  & 0.08 & Loc+Glob \\
& 70 rpm (R5)  & 0.09 & Min\_Vent & 0.07 & Global \\
& 100 rpm (R1) & 0.11 & Min\_Vent & 0.10 & Local \\
\midrule
\multirow{3}{*}{Stairs Climbing} 
& 60 Watts  & 0.12 & R\_Ankle\_ACCL     & 0.11 & Local+Global \\
& 75 Watts  & 0.11 & Min\_Vent  & 0.10 & Global \\
& 90 Watts  & 0.11 & Min\_Vent  & 0.11 & Hexoskin \\
\bottomrule
\end{tabular}%
}

\end{minipage}
\caption{NRMSE for different activities and different conditions.}
\label{tab:per_activ}
\end{table}

Second, performance varied with activity intensity. RMSEs were lower for low-intensity activities, while higher-intensity tasks produced larger RMSEs. However, Table \ref{tab:per_activ} showed that normalization (NRMSE) reduced these differences. Several high-intensity conditions (e.g., running at $1.8 \,\mathrm{m/s} $) also achieved comparable NRMSE. This indicates that while intensity increases error, models scale proportionally.

 Table \ref{tab:per_activ} also indicated that, as expected, minute ventilation emerged as the strongest single input and obtained the overall best NRMSE of $0.04$, while among grouped signals, Local+Global and Hexoskin consistently delivered the best performance across activities.

\subsection{Per-Subject Evaluation}

\begin{figure}[hbpt]
\centering

\begin{minipage}[t]{0.48\textwidth}
    \centering
    \includegraphics[width=\linewidth]{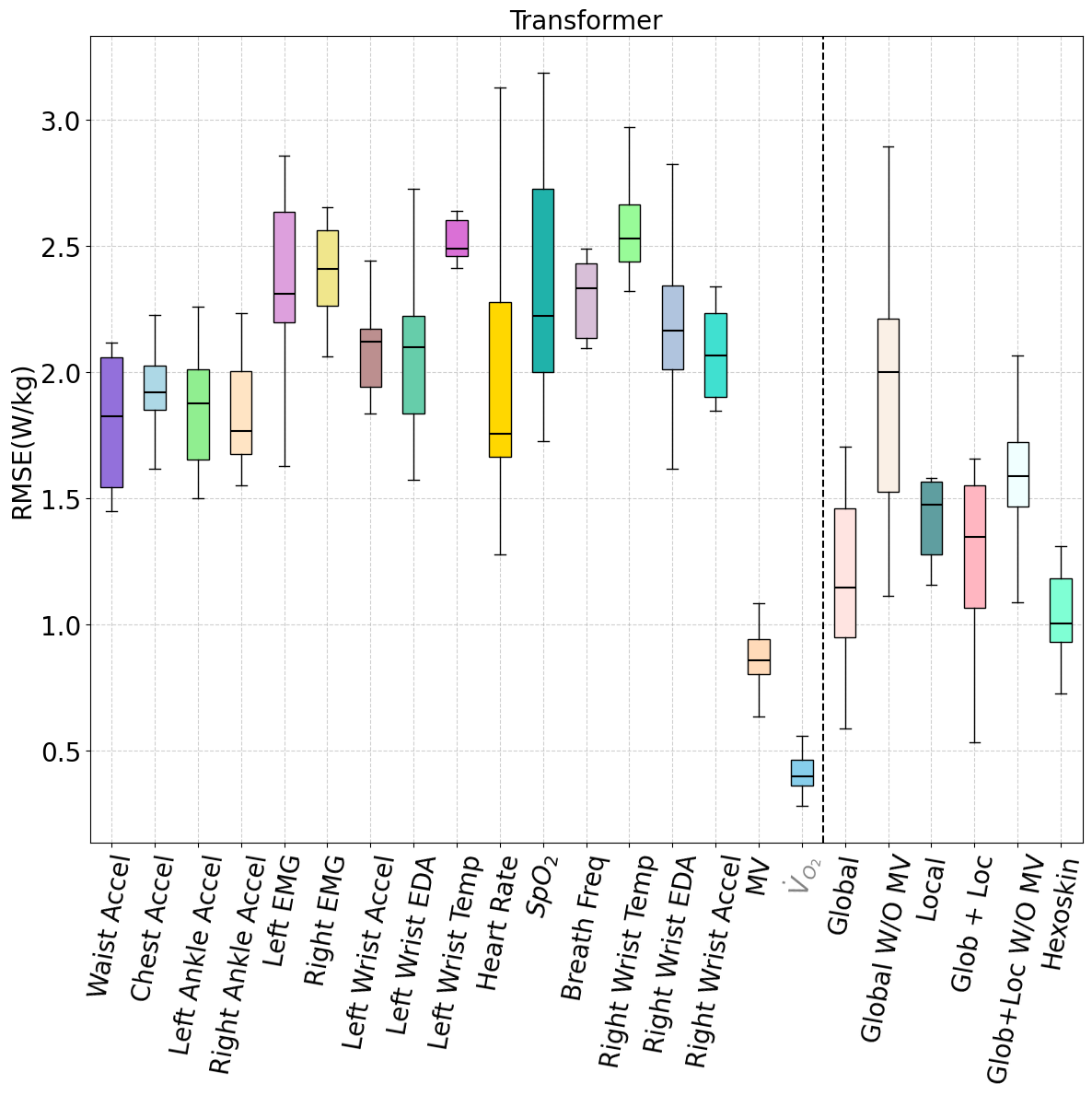}
\end{minipage}
\hfill
\begin{minipage}[t]{0.48\textwidth}
    \centering
    \includegraphics[width=\linewidth]{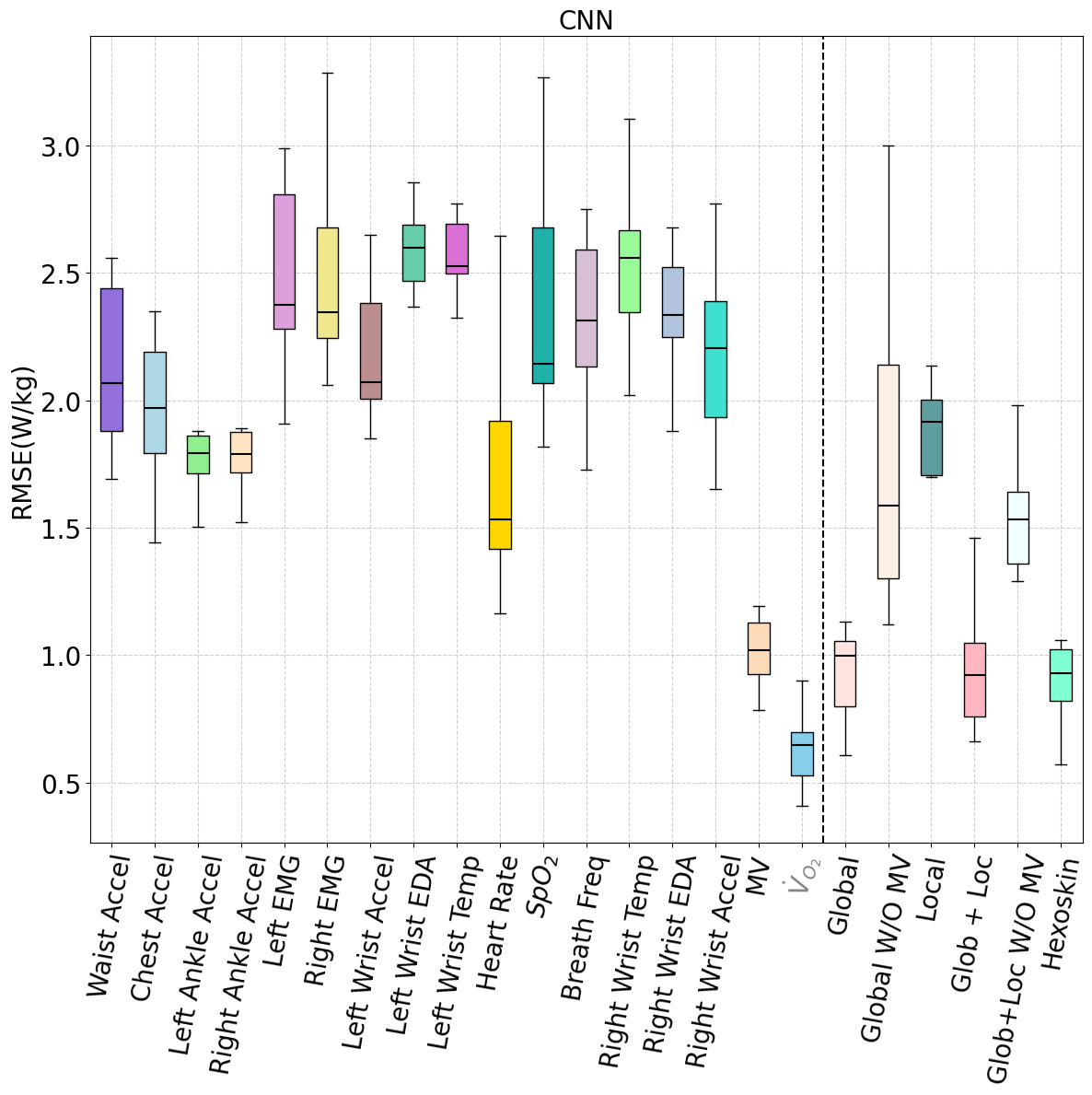}
\end{minipage}

\vspace{0.3cm}

\caption{Performance of Transformer and CNN for single and grouped signals across 10 subjects. The dashed line separates single- from grouped-signals in each plot. Boxplots represent the distribution of RMSE values across subjects: median (line), 25th–75th percentiles (box), and whiskers to 1.5×IQR. (MV: Minute Ventilation)}
\label{fig:transformer_vs_cnn}
\end{figure}

Lastly, we evaluated the effectiveness of different physiological signals and models per subject to examine how results fluctuate with individual differences. 
Figure \ref{fig:transformer_vs_cnn} compares CNN and Transformer models for single and grouped signals. While overall trends were consistent across architectures, we highlighted the most informative results here (further plots are available in the supplementary materials). Each boxplot shows the distribution of RMSE values across 10 subjects for a given input signal.
As expected, minute ventilation consistently yielded the lowest RMSE with minimal inter-subject variability, confirming its role as the most robust predictor of energy expenditure. In contrast, signals such as $SpO_{2}$ and EMG magnitude (left and right) showed both higher RMSE and greater variance, reflecting weak predictive power and strong inter-individual differences in signal quality. Heart rate achieved a low average RMSE but displayed high variance across subjects.

Interesting insights arose from comparing the variance differences between chest and ankle accelerations. 
Chest acceleration, which reflects global body motion, presumably benefited from the Transformer's ability to capture smooth, long-range dependencies, resulting in lower variance.
In contrast, CNNs, which rely on local temporal filters, may have failed to capture these patterns. On the other hand, ankle acceleration signals are periodic and structured patterns that were well-suited to the CNN's short-window convolutional architecture. Here, the Transformer may have been overly sensitive to small subject-specific gait variations, leading to higher variance. Finally, we observed that removing minute ventilation from the Global and Local+Global signal groups led to a noticeable increase in both median RMSE and variance.

\section{Discussion and Conclusion}

In this work, we implemented and systematically compared different neural network-based architectures for energy expenditure prediction from wearable physiological signals across diverse activities.
Our main objectives were to: (1) compare ML and DL models, (2) assess generalizability across signals and activities, and (3) analyze subject-specific variability.

\noindent \textbf{Model and signal configurations:} The Transformer and ResNet+Attention consistently outperformed other models, while CNNs offered a strong balance between accuracy and computational efficiency. Across all models, minute ventilation was the most reliable predictor, achieving an RMSE of $0.87 \,\mathrm{W/kg} $ with the Transformer. Since it is difficult to measure in practice, we examined alternatives: heart rate was the strongest single signal, and pairing or grouping signals further improved accuracy. For example, pairing heart rate with ankle acceleration or EMG signals across limbs reduced RMSE substantially. Signal fusion using Hexoskin and Local+Global inputs also outperformed single-signal baselines.

\noindent \textbf{Activity effects:} Performance varied with activity intensity. Low-intensity tasks (e.g., backward walking at $1 \,\mathrm{m/s} $) yielded very low errors, while higher-intensity activities showed greater RMSE. Normalization reduced these differences, with some intense activities achieving competitive NRMSE. This scaling of the error with intensity highlights the value of activity-specific refinements over universal models.

\noindent \textbf{Individual differences:} A central finding of this study is the extent of inter-subject variability. While minute ventilation provided stable performance for all participants, other signals such as heart rate and EMG were highly variable, likely reflecting physiological differences and variations in sensor quality. Model choice also interacted with signal type: Transformers captured smoother, whole-body dynamics (e.g., chest acceleration) more consistently across subjects, whereas CNNs better handled periodic patterns (e.g., ankle movement). These results highlight that robust EE estimation requires not only choosing the right signals but also matching model architecture to signal characteristics as well as individual variability.

\noindent \textbf{Further practical recommendations:} 
Taken together, our findings suggest several guidelines for real-world applications. When minute ventilation is available and processing time is less critical, the Transformer is the optimal choice. If faster inference is required and signals captured by the Hexoskin shirt are accessible, CNNs offer a good balance of efficiency and accuracy. In cases where both minute ventilation and EMG magnitude are available, ResNet+Attention provides the best overall accuracy. Finally, when minute ventilation cannot be measured, pairing heart rate with ankle acceleration and applying a CNN yields a strong and practical alternative.

Neural network–based approaches for EE prediction, particularly considering diverse physiological signals, have been understudied. Our results demonstrate both the potential of these methods and the substantial inter-subject variability that remains. This variability highlights the need for future work on activity-specific and personalized models. To encourage further research, we will release our code and models at \href{https://github.com/Sarvibabakhani/deeplearning-biosignals-ee}{this GitHub repository}.

\section{Acknowledgment}
The project was funded by Deutsche Forschungsgemeinschaft (DFG, German Research Foundation) under Germany’s Excellence Strategy (EXC 2075 - 390740016). We acknowledge the support of the Stuttgart Center for Simulation Science (SimTech) and the International Max Planck Research School for Intelligent Systems (IMPRS-IS).
The authors gratefully acknowledge the computing time provided on the high-performance computer HoreKa by the National High-Performance Computing Center at KIT (NHR@KIT). This center is jointly supported by the Federal Ministry of Education and Research and the Ministry of Science, Research and the Arts of Baden-Württemberg, as part of the National High-Performance Computing (NHR) joint funding program (https://www.nhr-verein.de/en/our-partners). HoreKa is partly funded by the German Research Foundation (DFG).
\bibliography{egbib}

\clearpage
\appendix

\renewcommand{\thesection}{S\arabic{section}}
\setcounter{section}{0}
\setcounter{figure}{0}
\setcounter{table}{0}
\renewcommand{\thefigure}{S\arabic{figure}}
\renewcommand{\thetable}{S\arabic{table}}


\pagestyle{bmvcsup}
\thispagestyle{bmvcsup} 
\input{appendix.tex}

\end{document}

%% file: appendix.tex
\markboth{BABAKHANI ET AL}{Supplementary Material} 
\section*{Supplementary Material}

\addcontentsline{toc}{section}{Supplementary Material} 

\vspace{0.5em}
\hrule
\vspace{1em}

\section{Further Dataset Information}
The dataset used in this study includes recordings from 10 subjects, each performing six types of physical activities: walking, running, incline walking, backward walking, cycling, and stair climbing. These activities were divided into two main experimental sessions:\\
Session 1: Sitting, standing, level walking, incline walking, and backward walking on a treadmill\\
Session 2: Sitting, standing, running on a treadmill, cycling on a stationary bike, and stair climbing on a stairmill.\\
For each activity, subjects followed a protocol in which they stood quietly for 6 minutes, performed each speed/resistance condition for 6 minutes in randomized order, and then sat quietly for another 6 minutes. All sensor data were recorded with synchronized timestamps, and each time step was manually annotated with the corresponding activity type.

\noindent \textbf{Below, we mentioned the utilized signals:}\\   
(i) acceleration magnitudes from the left/right ankles and wrist, waist, and chest
(For all acceleration signals, the vector magnitude across the three axes was calculated).\\
(ii) right/left wrist electrodermal activity (EDA) and skin temperature\\
(iii) a composite lower-limb signal derived from normalized sEMG envelopes\\
(iv) respiratory and cardiovascular measures, including $\dot{V}_{O_{2}}$, $\dot{V}_{CO_{2}}$, $SpO_{2}$, breath frequency, minute ventilation, and heart rate. Respiratory measures were collected breath-by-breath with a portable respirometer, while heart rate and $SpO_{2}$ were measured with a chest strap and earlobe oximeter. All signals were synchronized with the respirometer and stored on a breath-by-breath basis. Because these signals were measured breath-by-breath, their sampling frequency varied. We maintained a constant sample rate by averaging over the frequency of breath for each activity and its specific condition (e.g., backward walking at $1 \,\mathrm{m/s} $ vs. $0.7\,\mathrm{m/s} $) \cite{Robergs}.  \\
\textbf{The ground truth energy expenditure:} It was computed using the Brockway equation~\cite{Brockway}, which relies on measurements of $\dot{V}_{O_{2}}$a and $\dot{V}_{CO_{2}}$. The resulting values were normalized to body weight for comparability across subjects. Steady-state EE was estimated by averaging the final three minutes of each six-minute activity condition. To obtain the net energetic cost, the standing baseline value recorded at the start of each trial was subtracted from the steady-state estimate.
(Further details on data collection and processing can be found in \cite{K.A.Ingraham}.)\\
\textbf{Input formatting and fusion:} In all experiments, input signals were segmented into fixed-length windows of 10 or 20 time steps. The final choice of time step size was selected based on preliminary performance tuning. For multi-signal inputs, we applied early fusion by concatenating the signals along the feature dimension.

\section{Detailed Model Architecture and Training Procedure}
In this study, we tested six models: Linear Regression, CNN, LSTM, ResNet, ResNet+Attention, and Transformer. For each model, we provide details on the network architecture, training configuration, and implementation choices, including layer design and optimization settings.
\subsection{Linear Regression:}
We implemented both single and multiple linear regression models for the energy expenditure (EE) estimation. The general form of the model is:
\begin{equation}
\hat{y} = b_0 + \sum_{i=1}^{n} b_i x_i = Xb
\end{equation}
where $\hat{y}$ represents the vector of predicted EE values, the variable $n$ denotes the number of input signals included in the model.
The input matrix $X$ consists of a column of ones to account for the bias term and $n$ columns representing the input signals. The vector $b$ contains the learned regression coefficients. 
\subsection{CNN:}
The CNN model consists of three 1D convolutional blocks followed by fully connected layers. Each convolutional block includes a 1D convolution layer (kernel size = 3), batch normalization, ReLU activation, max pooling (kernel size = 2), with dropout applied in the second and third blocks.
The number of filters decreases across the layers (64, 32, and 16).\\
The convolutional output is flattened and passed through two fully connected layers:
The first is a dense layer with 40 units, ReLU activation, and dropout.
The second is an output layer with linear activation to match the target dimension.
We used 20 time steps, a batch size of 8, and the Adam optimizer (learning rate = 0.0005) for training this network.

\subsection{LSTM:}
We implemented a stacked LSTM-based regression network. The model consists of two sequential Long Short-Term Memory (LSTM) layers. The first LSTM layer has 128 hidden units, followed by dropout regularization. Its output is passed to a second LSTM layer with 64 hidden units and an additional dropout. The final LSTM output is flattened and passed through a fully connected layer with 64 units, followed by batch normalization and dropout.
We set the number of time steps to 20, used a batch size of 32, and used the Adam optimizer with a learning rate of 0.0005.

\subsection{ResNet:}
The original ResNet architecture is adapted for 1D time-series input.
The model architecture begins with a 1D convolution using 64 filters (kernel size = 7), followed by batch normalization, ReLU activation, and max pooling.
Next, there are three residual blocks with increasing output dimensions
(64 to 128 channels, 128 to 256 channels, and 256 to 512 channels).
Each block contains two Conv1D layers (kernel size = 3) with batch normalization layers and skip connections (including a convolution to match the input and output dimensions). After the final residual block, global average pooling is applied over the time dimension, followed by a linear layer mapping the pooled features to the desired output size.
The network was trained with 10 time steps, a batch size of 32, and the Adam optimizer with a learning rate of 0.001.

\subsection{ResNet+Attention:}
In this architecture, there is an attention block added to the residual blocks in our ResNet architecture. This block computes a self-attention mechanism over the temporal dimension. Within this block, three distinct 1×1 convolutions are applied to derive the query, key, and value representations of the input.

The attention score computed using the attention function equation from \cite{vaswani}, based on a scaled dot-product attention mechanism over the temporal dimension.
After re-weighting the value features, a residual connection integrates the attention output with the original input, ensuring that the initial features are preserved. The network was trained with 10 time steps, a batch size of 8, and the Adam optimizer (learning rate = 0.0005).

\subsection{Transformer:}
This model is based on the Transformer encoder framework \cite{vaswani}, adapted for sequential signal modeling. The raw input signal is first projected into a higher-dimensional representation using a 1D convolution with kernel size 3, which captures local temporal patterns. Since the Transformer architecture does not contain recurrence or convolutional structure, temporal order is incorporated through sinusoidal positional encodings as introduced in \cite{vaswani}. The projected sequence is then processed by a stack of two Transformer encoder layers, each consisting of 8-head self-attention, a feedforward network with hidden dimension 256, residual connections, and layer normalization. Finally, the encoded sequence is passed through a lightweight feedforward output head composed of two fully connected layers with a ReLU activation in between to produce predictions at each time step. For training, we used a time step length of 10, a batch size of 4, and the Adam optimizer with a learning rate of 0.0009.

\section{Additional Per-activity Evaluation Tables}
In "Per-Activity Evaluation" (Section 3.4) of the paper, we discussed how model performance varied across activities. To complement that analysis, we provide a detailed summary of pairwise signal combinations here. Table \ref{tab:bestworst_pairs} presents the worst-performing pairs (left) and the best partner for each signal when minute ventilation was excluded (right). These results highlight which modalities provide complementary information and which pairs lead to consistently poor predictions.
\begin{table}[hbpt]
\centering
\begin{minipage}[t]{0.49\textwidth}
\centering
\scalebox{0.65}{ 
  \renewcommand{\arraystretch}{1.2}
\begin{tabular}{l|l|l|c}
\hline
\textbf{Signal 1} & \textbf{Signal 2} & \textbf{Model} & \textbf{RMSE (W/kg))} \\
\hline
L\_Wrist\_Temp & R\_Wrist\_Temp & ResNet & 3.10 \\
Waist\_ACCL & L\_Wrist\_ACCL & ResAtt & 3.12 \\
R\_Wrist\_Elec & R\_Wrist\_Temp & ResNet & 3.15 \\
L\_Wrist\_Elec & L\_Wrist\_Temp & Lin-Reg & 3.19 \\
L\_Wrist\_Temp & R\_Wrist\_Temp & Trans & 3.21 \\
L\_Wrist\_Elec & R\_Wrist\_Temp & Lin-Reg & 3.23 \\
L\_Wrist\_Temp & R\_Wrist\_Temp & LSTM & 3.25 \\
EMG\_M\_L & SpO2 & CNN & 3.35 \\
Waist\_ACCL & Chest\_ACCL & ResAtt & 3.36 \\
EMG\_M\_R & $SpO_{2}$ & CNN & 3.44 \\
Waist\_ACCL & R\_Wrist\_ACCL & ResAtt & 3.81 \\
EMG\_M\_R & $SpO_{2}$ & LSTM & 4.36 \\
EMG\_M\_L & R\_Wrist\_Elec & CNN & 4.51 \\
EMG\_M\_R & R\_Wrist\_Elec & CNN & 5.28 \\
EMG\_M\_R & L\_Wrist\_Elec & CNN & 7.90 \\
EMG\_M\_L & L\_Wrist\_Elec & CNN & 8.05 \\
\hline
\end{tabular}%
}
\end{minipage}
\hspace{0.01\textwidth}%
\begin{minipage}[t]{0.49\textwidth}
\centering
\scalebox{0.65}{ 
  \renewcommand{\arraystretch}{1.2}
\begin{tabular}{l|l|l|c}
\hline
\textbf{Signal} & \textbf{Best Pair} & \textbf{Model} & \textbf{RMSE (W/kg))} \\
\hline
Waist\_ACCL & EMG\_M\_L & Trans & 1.64 \\
Chest\_ACCL & HR & CNN & 1.67 \\
L\_Ankle\_ACCL & HR & CNN & 1.51 \\
R\_Ankle\_ACCL & HR & CNN & 1.49 \\
L\_Wrist\_ACCL & HR & CNN & 1.79 \\
L\_Wrist\_Elec & HR & CNN & 1.63 \\
L\_Wrist\_Temp & R\_Ankle\_ACCL & CNN & 1.93 \\
R\_Wrist\_Elec & HR & CNN & 1.64 \\
R\_Wrist\_Temp & R\_Ankle\_ACCL & CNN & 1.87 \\
R\_Wrist\_ACCL & HR & CNN & 1.81 \\
EMG\_M\_L & Waist\_ACCL & Trans & 1.64 \\
EMG\_M\_R & L\_Ankle\_ACCL & ResAtt & 1.53 \\
HR & R\_Ankle\_ACCL & CNN & 1.49 \\
$SpO_{2}$ & HR & CNN & 1.86 \\
Breath\_Freq & R\_Ankle\_ACCL & CNN & 1.91 \\
Min\_Vent & EMG\_M\_L & ResAtt & 0.90 \\
\hline
\end{tabular}%
}
\end{minipage}
\caption{\textbf{Left}: Worst signal pair combinations. \textbf{Right}: Best pair for each signal (in absence of Minute Ventilation).}
\label{tab:bestworst_pairs}
\end{table}

\section{Additional Per-Subject Evaluation Plots}
\begin{figure}[hbpt]
    \centering

    \begin{minipage}[t]{0.48\textwidth}
        \centering
        \includegraphics[width=\linewidth]{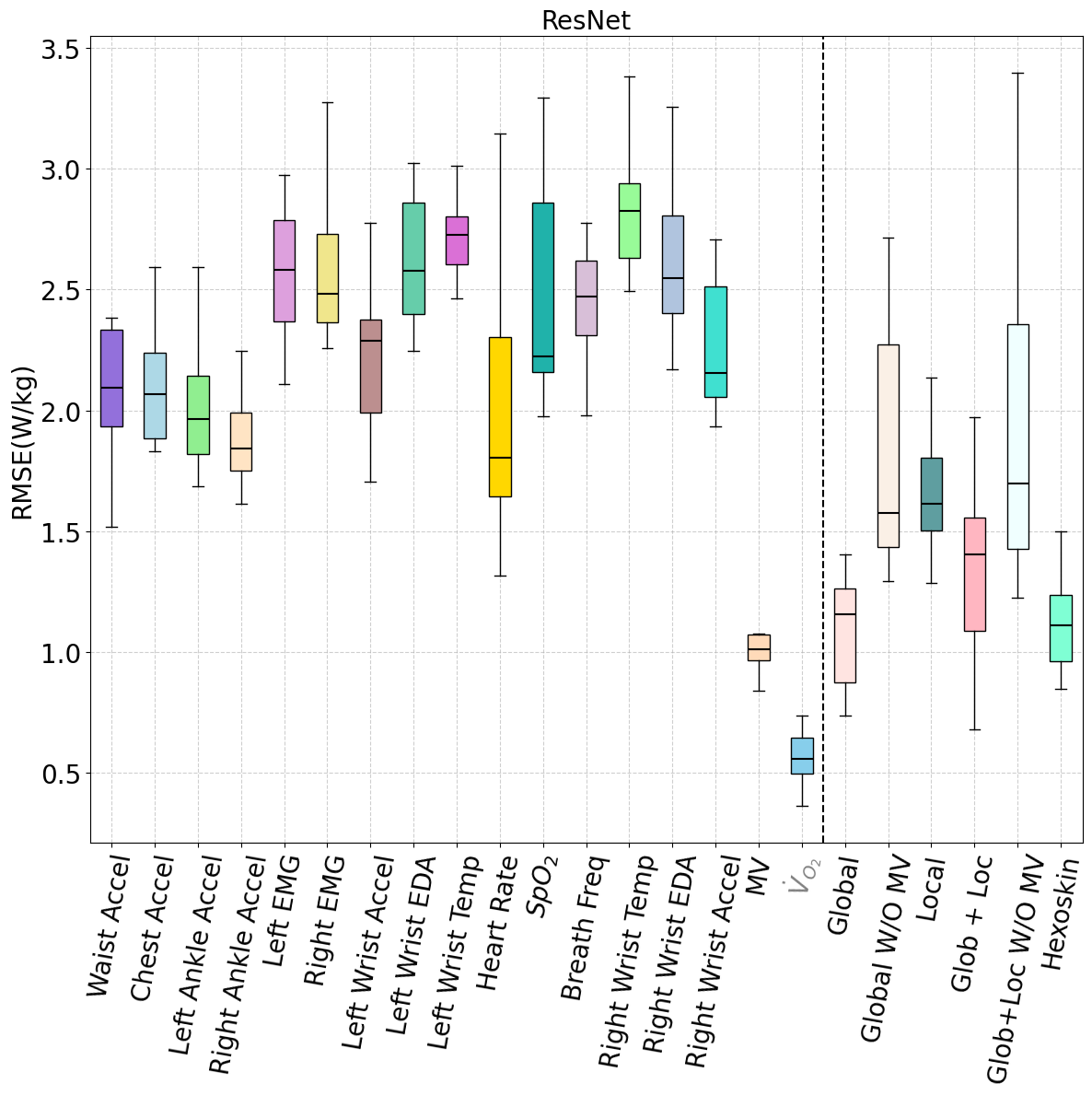}
    \end{minipage}
    \hfill
    \begin{minipage}[t]{0.48\textwidth}
        \centering
        \includegraphics[width=\linewidth]{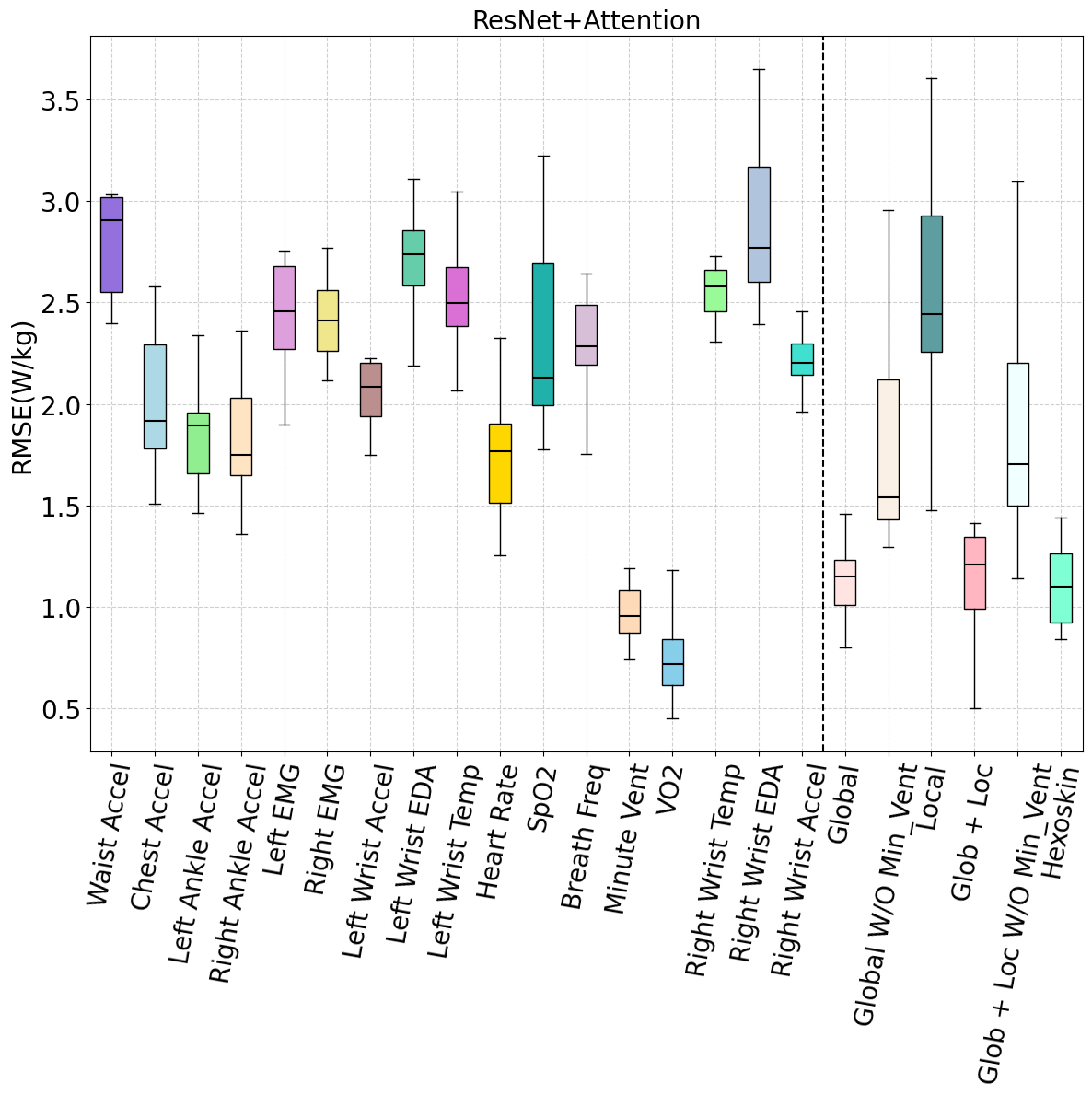}
    \end{minipage}

    \vspace{0.3cm}

    \begin{minipage}[t]{0.48\textwidth}
        \centering
        \includegraphics[width=\linewidth]{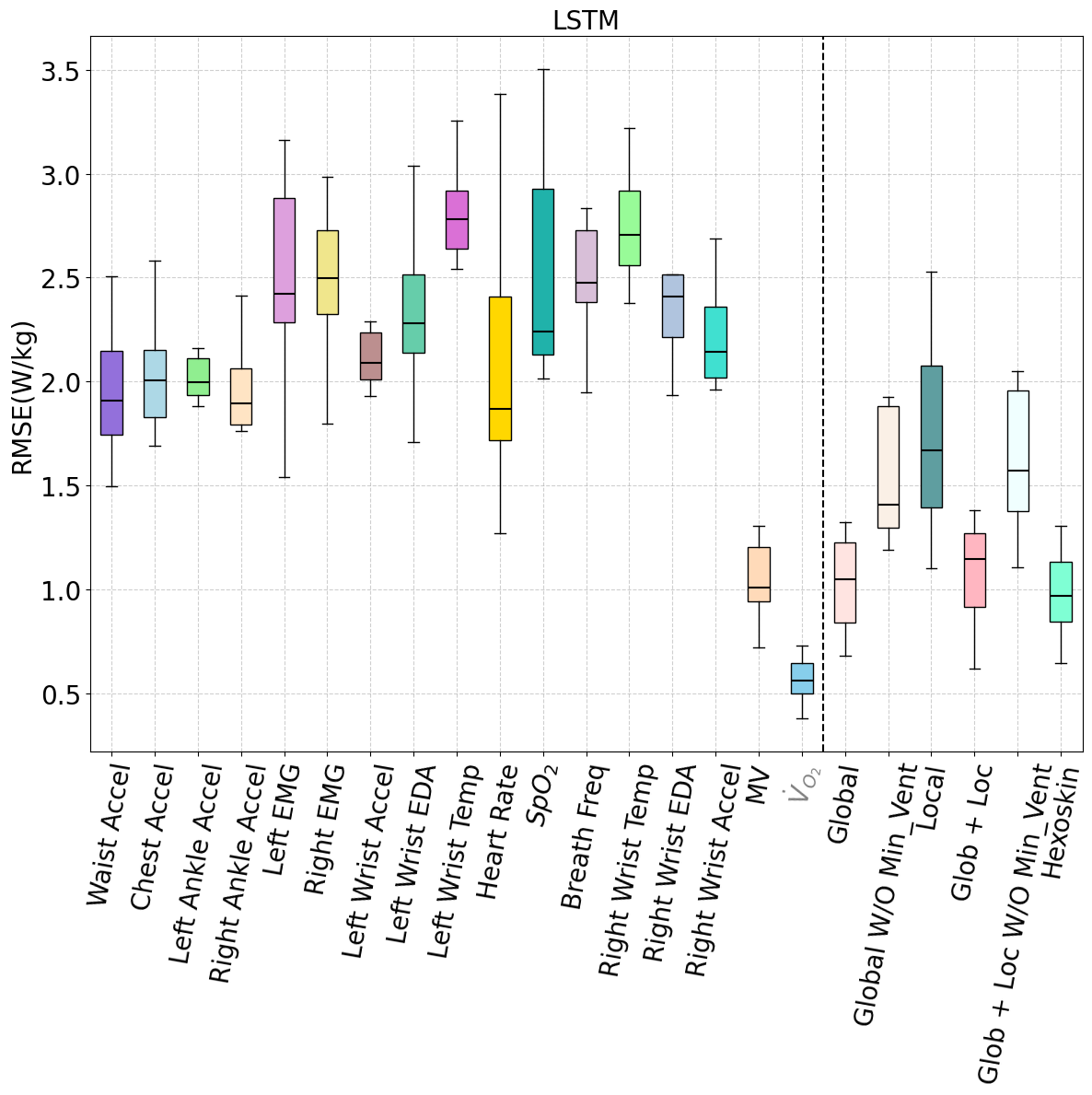}
    \end{minipage}
    \hfill
    \begin{minipage}[t]{0.48\textwidth}
        \centering
        \includegraphics[width=\linewidth]{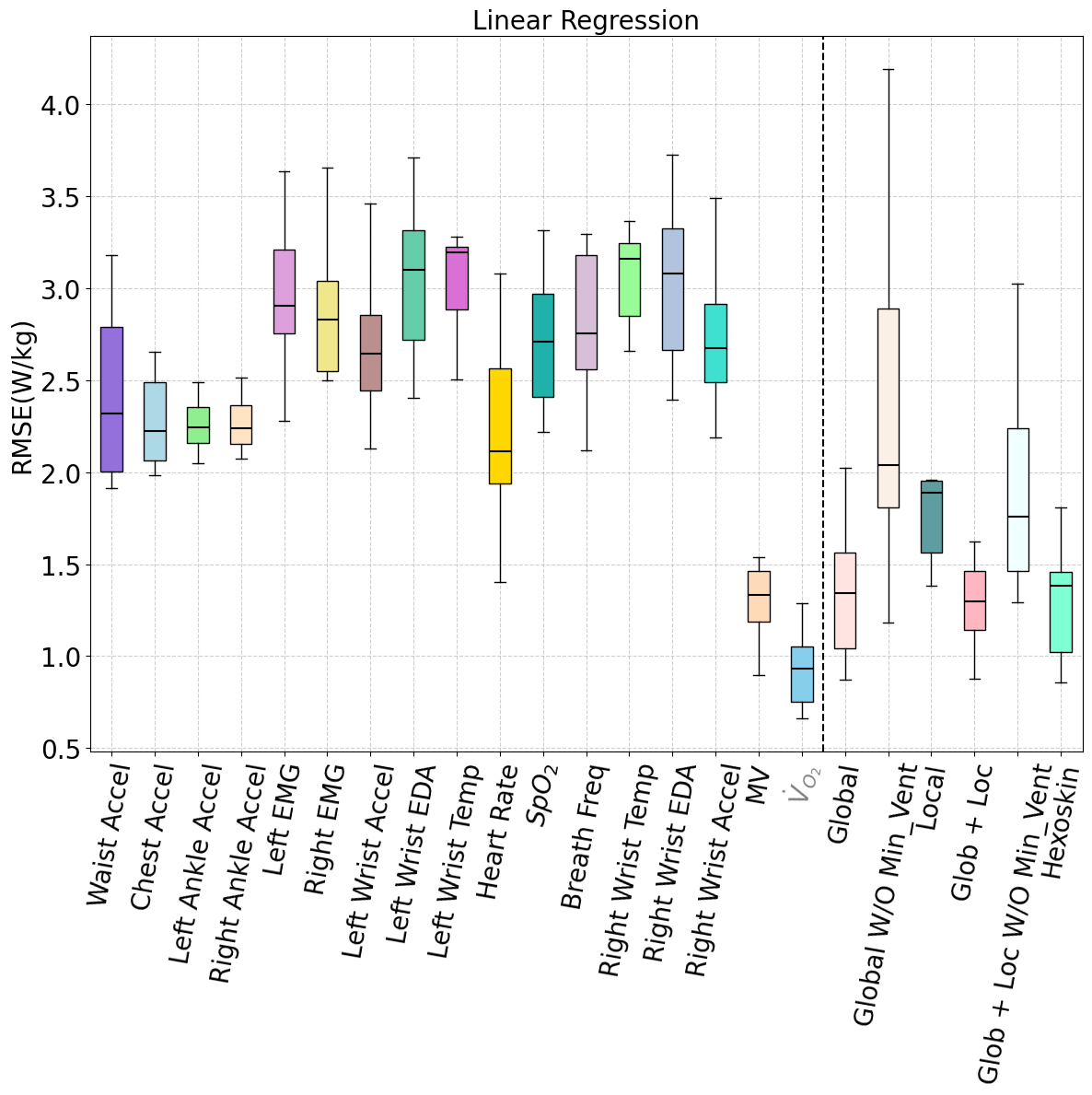}
    \end{minipage}
\vspace{0.3cm}
    \caption{Performance of the remaining four models for single and grouped signals across 10 subjects (complementing Section 3.5 of the main paper). The dashed line separates single- from grouped-signals in each plot. Boxplots represent the distribution of RMSE values across subjects: median (line), 25th–75th percentiles (box), and whiskers to 1.5×IQR. (MV: Minute Ventilation)}
    \label{fig:6grid}
\end{figure}

In "Per-Subject Evaluation" (Section 3.5) of the paper, we focused on the Transformer and CNN models and their corresponding plots. In Figure \ref{fig:6grid}, we provide the per-subject performance plots for the four remaining models. The trends largely mirror those observed for the Transformer and CNN models, confirming the key observations in the main paper.

%% file: egbib.bib
@article{K.A.Ingraham,
	author  = {Ingraham, Kimberly A. and Ferris, Daniel P. and Remy, C. David},
	title   = {Evaluating physiological signal salience for estimating metabolic energy cost from wearable sensors},
	journal = {Journal of Applied Physiology},
	year    = {2019},
	volume  = {126},
	number  = 3,
	pages   = {717--729},
        doi     = {10.1152/japplphysiol.00714.2018}
}

@inproceedings{xu2024spatial,
  author    = {Xu, Wenjin and Meng, Wei and Zhu, Chang and Kong, Jingjing and Liu, Quan and Ai, Qingsong},
  title     = {Spatial-Temporal Fusion Network with Hybrid Attention for Energy Expenditure Prediction Based on Multi-Sensor},
  booktitle = {Proceedings of the 2024 30th International Conference on Mechatronics and Machine Vision in Practice (M2VIP)},
  year      = {2024},
  pages     = {1--6},
  doi       = {10.1109/M2VIP62491.2024.10746155}
}

@article{P.Slade,
  author  = {Slade, Patrick and Atkeson, Christopher and Donelan, J. Maxwell and Houdijk, Han and Ingraham, Kimberly A. and Kim, Myunghee and Kong, Kyoungchul and Poggensee, Katherine L. and Riener, Robert and Steinert, Martin and Zhang, Juanjuan and Collins, Steven H.},
  title   = {On human-in-the-loop optimization of human–robot interaction},
  journal = {Nature},
  year    = {2024},
  volume  = {633},
  number  = {8031},
  pages   = {779--788},
  doi     = {10.1038/s41586-024-07697-2},
}

@inproceedings{Koller2016Body,
  author    = {Koller, Jeffrey R. and Gates, Deanna H. and Ferris, Daniel P. and Remy, C. David},
  title     = {‘Body-in-the-Loop’ Optimization of Assistive Robotic Devices: A Validation Study},
  booktitle = {Proceedings of Robotics: Science and Systems (RSS) XII},
  year      = {2016},
  address   = {Ann Arbor, MI, USA},
  pages     = {1--10},
}

@article{felt2015body,
  author  = {Felt, Wyatt and Selinger, Jessica C. and Donelan, J. Maxwell and Remy, C. David},
  title   = {“Body-In-The-Loop”: Optimizing Device Parameters Using Measures of Instantaneous Energetic Cost},
  journal = {PLoS ONE},
  year    = {2015},
  volume  = {10},
  number  = {8},
  pages   = {e0135342},
  doi     = {10.1371/journal.pone.0135342},
}

@article{Lee,
author = {Lee, Chang June and Lee, Jung},
year = {2024},
month = {01},
pages = {414},
title = {IMU-Based Energy Expenditure Estimation for Various Walking Conditions Using a Hybrid CNN–LSTM Model},
volume = {24},
journal = {Sensors},
doi = {10.3390/s24020414}
}

@article{Slade2,
	author  = {Slade, Patrick and Kochenderfer, Mykel J. and Delp, Scott L. and Collins, Steven H.},
	title   = {Sensing leg movement enhances wearable monitoring of energy expenditure},
	journal = {Nature Communications},
	year    = {2021},
	volume  = {12},
	number  = 1,
	pages   = {4312},
        doi     = {10.1038/s41467-021-24173-x}
}

@article{Yuan,
	author  = {Yuan, Jinfeng and Zhang, Yuzhong and Liu, Shiqiang and Zhu, Rong},
	title   = {Wearable leg movement monitoring system for high-precision real-time metabolic energy estimation and motion recognition},
	journal = {Research},
	year    = {2023},
	volume  = {6},
	pages   = {0214},
        doi     = {10.34133/research.0214}
}

@article{Lopes,
	author  = {Lopes, João M. and Figueiredo, Joana and Fonseca, Pedro and Cerqueira, João J. and Vilas-Boas, João P. and Santos, Cristina P.},
	title   = {Deep learning-based energy expenditure estimation in assisted and non-assisted gait using inertial, EMG, and heart rate wearable sensors},
	journal = {Sensors},
	year    = {2022},
	volume  = {22},
	number  = 20,
	pages   = {7913},
        doi     = {10.3390/s22207913}
}

@article{Slade3,
	author  = {Slade, Patrick and Troutman, Rachel and Kochenderfer, Mykel J. and Collins, Steven H. and Delp, Scott L.},
	title   = {Rapid energy expenditure estimation for ankle assisted and inclined loaded walking},
	journal = {Journal of NeuroEngineering and Rehabilitation},
	year    = {2019},
	volume  = {16},
	number  = 1,
	pages   = {67},
        doi     = {10.1186/s12984-019-0535-7}
}

@article{Falcone,
	author  = {Falcone, Tiziana and Del Ferraro, Simona and Molinaro, Vincenzo and Zollo, Loredana and Lenzuni, Paolo},
	title   = {Estimation of the metabolic rate in the occupational field: a regression model using accelerometers},
	journal = {International Journal of Industrial Ergonomics},
	year    = {2023},
	volume  = {96},
	pages   = {103454},
        doi     = {10.1016/j.ergon.2023.103454}
}

@inproceedings{Marena,
	author    = {Marena, Marco and Ratnakumar, Neethan and Jones, Rachel and Zhou, Xianlian and Das, Sanchoy and Shen, Bo},
	title     = {Predicting metabolic rate for firefighting activities with worn loads using a heart rate sensor and machine learning},
	booktitle = {Proceedings of the IEEE International Conference on Body Sensor Networks (BSN)},
	pages     = {1--4},
	publisher = {IEEE},
	address   = {Boston, MA, USA},
	year      = {2023},
	doi       = {10.1109/BSN58485.2023.10331063}
}

@inproceedings{Monteiro2,
	author    = {Monteiro, Sara and Figueiredo, Joana and Santos, Cristina},
	title     = {Towards a more efficient human-exoskeleton assistance},
	booktitle = {Proceedings of the IEEE International Conference on Autonomous Robot Systems and Competitions (ICARSC)},
	pages     = {181--186},
	publisher = {IEEE},
	address   = {Tomar, Portugal,},
	year      = {2023},
	doi       = {10.1109/ICARSC58346.2023.10129556}
}

@article{monteiro2024hitl,
  author  = {Monteiro, Sara and Figueiredo, Joana and  Fonseca, Pedro and Vilas-Boas, J. Paulo  and Santos, Cristina P.},
  title   = {Human-in-the-Loop Optimization of Knee Exoskeleton Assistance for Minimizing User’s Metabolic and Muscular Effort},
  journal = {Sensors},
  year    = {2024},
  volume  = {24},
  number  = {11},
  pages   = {3305},
  doi     = {10.3390/s24113305}
}

@article{houssein2023energy,
  author  = {Houssein, Aya and Prioux, Jacques and  Gastinger, Steven and  Martin, Brice and  Zhou, Fenfen and Ge, Di},
  title   = {Energy Expenditure Estimation From Respiratory Magnetometer Plethysmography: A Comparison Study},
  journal = {IEEE Journal of Biomedical and Health Informatics},
  year    = {2023},
  volume  = {27},
  number  = {5},
  pages   = {2345--2352},
  doi     = {10.1109/JBHI.2023.3252173}
}

@article{Ni,
	author  = {Ni, Zhiqiang and Wu, Tongde and Wang, Tao and Sun, Fangmin and Li, Ye},
	title   = {Deep multi-branch two-stage regression network for accurate energy expenditure estimation with ECG and IMU data},
	journal = {IEEE Transactions on Biomedical Engineering},
	year    = {2022},
	volume  = {69},
	number  = {10},
	pages   = {3224--3233},
	doi     = {10.1109/TBME.2022.3163429}
}

@article{Cvetković,
	author  = {Cvetković, Bozidara and Milić, Radoje and Luštrek, Mitja},
	title   = {Estimating energy expenditure with multiple models using different wearable sensors},
	journal = {IEEE Journal of Biomedical and Health Informatics},
	year    = {2015},
	volume  = {19},
	number  = {5},
	pages   = {1574--1581},
	doi     = {10.1109/JBHI.2015.2432911}
}

@article{Robergs,
	author  = {Robergs, R.A. and Dwyer, D. and Astorino, T.},
	title   = {Recommendations for improved data processing from expired gas analysis indirect calorimetry},
	journal = {Sports Medicine},
	year    = {2010},
	volume  = {40},
	number  = {2},
	pages   = {95--111}
}

@article{Brockway,
	author  = {Brockway, J.M.},
	title   = {Derivation of formulae used to calculate energy expenditure in man},
	journal = {Human Nutrition: Clinical Nutrition},
	year    = {1987},
	volume  = {41},
	number  = {6},
	pages   = {463--471}
}

@inproceedings{vaswani,
	author    = {Vaswani, Ashish and Shazeer, Noam and Parmar, Niki and Uszkoreit, Jakob and Jones, Llion and Gomez, Aidan N. and Kaiser, Lukasz and Polosukhin, Illia},
	title     = {Attention is All You Need},
	booktitle = {Advances in Neural Information Processing Systems (NeurIPS)},
	volume    = {30},
	pages     = {5998--6008},
	publisher = {Curran Associates, Inc.},
	address   = {Long Beach, CA, USA},
	year      = {2017}
}

@article{resi,
	author = {He, Kaiming and Zhang, Xiangyu and Ren, Shaoqing and Sun, Jian},
	title = {Deep Residual Learning for Image Recognition},
	journal = {Proceedings of the IEEE Conference on Computer Vision and Pattern Recognition (CVPR)},
	year = {2016},
	volume = {2016},
	number = {1},
	pages = {770-778}}

@misc{hexo,
  author       = {{Carr\'e Technologies Inc. (Hexoskin)}},
  title        = {Hexoskin Smart Shirts},
  year         = {2025},
  howpublished = {\url{https://hexoskin.com/}},
  
}

@article{masullo2018calorinet,
  title={CaloriNet: From silhouettes to calorie estimation in private environments},
  author={Masullo, Alessandro and Burghardt, Tilo and Damen, Dima and Hannuna, Sion and Ponce-L{\'o}pez, V{\'\i}ctor and Mirmehdi, Majid},
  journal={British Machine Vision Conference (BMVC)},
  year={2018}
}

@inproceedings{peng2022should,
  title={Should I take a walk? Estimating energy expenditure from video data},
  author={Peng, Kunyu and Roitberg, Alina and Yang, Kailun and Zhang, Jiaming and Stiefelhagen, Rainer},
  booktitle={Proceedings of the IEEE/CVF Conference on Computer Vision and Pattern Recognition},
  pages={2075--2085},
  year={2022}
}

@inproceedings{kasturi2024estimating,
  title={Estimating Physical Activity Energy Expenditure from Video},
  author={Kasturi, Gayatri and Shrestha, Pragya and Strath, Scott J and Kate, Rohit J},
  booktitle={2024 IEEE EMBS International Conference on Biomedical and Health Informatics (BHI)},
  pages={1--8},
  year={2024},
  organization={IEEE}
}

@article{kim,
AUTHOR = {Kim, Min-Seo and Seong, Ju-Hyeon},
TITLE = {A Personalized Energy Expenditure Estimation Method Using Modified MET and Heart Rate-Based DQN},
JOURNAL = {Sensors},
VOLUME = {25},
YEAR = {2025},
NUMBER = {11},
ARTICLE-NUMBER = {3416},
URL = {https://www.mdpi.com/1424-8220/25/11/3416},
ISSN = {1424-8220},
DOI = {10.3390/s25113416}
}
